\documentclass[letterpaper, 10 pt,twocolumn, conference]{ieeeconf}

\IEEEoverridecommandlockouts                    

\usepackage[ruled]{algorithm2e}
\usepackage{amsmath} 
\usepackage{amssymb}  
\usepackage{amsfonts}  
\usepackage{graphicx} 
\usepackage{booktabs}
\usepackage[normalem]{ulem}
\useunder{\uline}{\ul}{}
\usepackage[caption=false, font=footnotesize]{subfig}
\usepackage{cuted}
\usepackage{microtype}
\usepackage{fancyhdr}
\usepackage{balance}
\usepackage[colorlinks=true,
        linkcolor=black,   
        citecolor=black,   
        filecolor=black,
        urlcolor=black,
        bookmarks=true,
        bookmarksopen=true,
        bookmarksopenlevel=3,
        plainpages=false,
        pdfpagelabels=true]{hyperref}
\usepackage[all]{hypcap}
\usepackage{microtype}
\usepackage{cite}
\usepackage{booktabs}

\graphicspath{ {images/} } 

\usepackage{xcolor}

\newcommand{\vect}[1]{\mathbf{ #1}}

\newcommand{\vectg}[1]{{\boldsymbol{ #1}}}

\newcommand{\argmin}{\operatornamewithlimits{argmin}}

\newcommand{\argmax}{\operatornamewithlimits{argmax}}

\newcommand{\va}{\vect{a}}

\newcommand{\vs}{\vect{s}}

\newcommand{\vu}{\vect{u}}

\newcommand{\vpi}{\vectg{\pi}}

\newcommand{\vtheta}{\vectg{\theta}}

\title{\LARGE \bf
What Would You Do? Acting by Learning to Predict
}

\author{Adam W.~Tow, Niko S\"underhauf, Sareh Shirazi, Michael Milford, J\"urgen Leitner
\thanks{The authors are with the Australian Centre for Robotic Vision (ACRV) at the Queensland University of Technology (QUT), Brisbane, Australia.}%
\thanks{EMail: {\tt\small adam.tow@qut.edu.au}}
\thanks{This research was supported by the Australian Research Council Centre of Excellence for Robotic Vision (ACRV) (project number CE140100016).}
}

\begin{document}

\maketitle


\begin{abstract}




We propose to learn tasks directly from visual demonstrations by learning to predict the \textit{outcome} of human and robot actions on an environment. We enable a robot to physically perform a human demonstrated task without knowledge of the thought processes or actions of the human, only their visually observable state transitions. We evaluate our approach on two table-top, object manipulation tasks and demonstrate generalisation to previously unseen states. Our approach reduces the priors required to implement a robot task learning system compared with the existing approaches of Learning from Demonstration, Reinforcement Learning and Inverse Reinforcement Learning. 

\end{abstract}

\section{Introduction}





\begin{figure}[t]
    \centering
    \includegraphics[width=\linewidth]{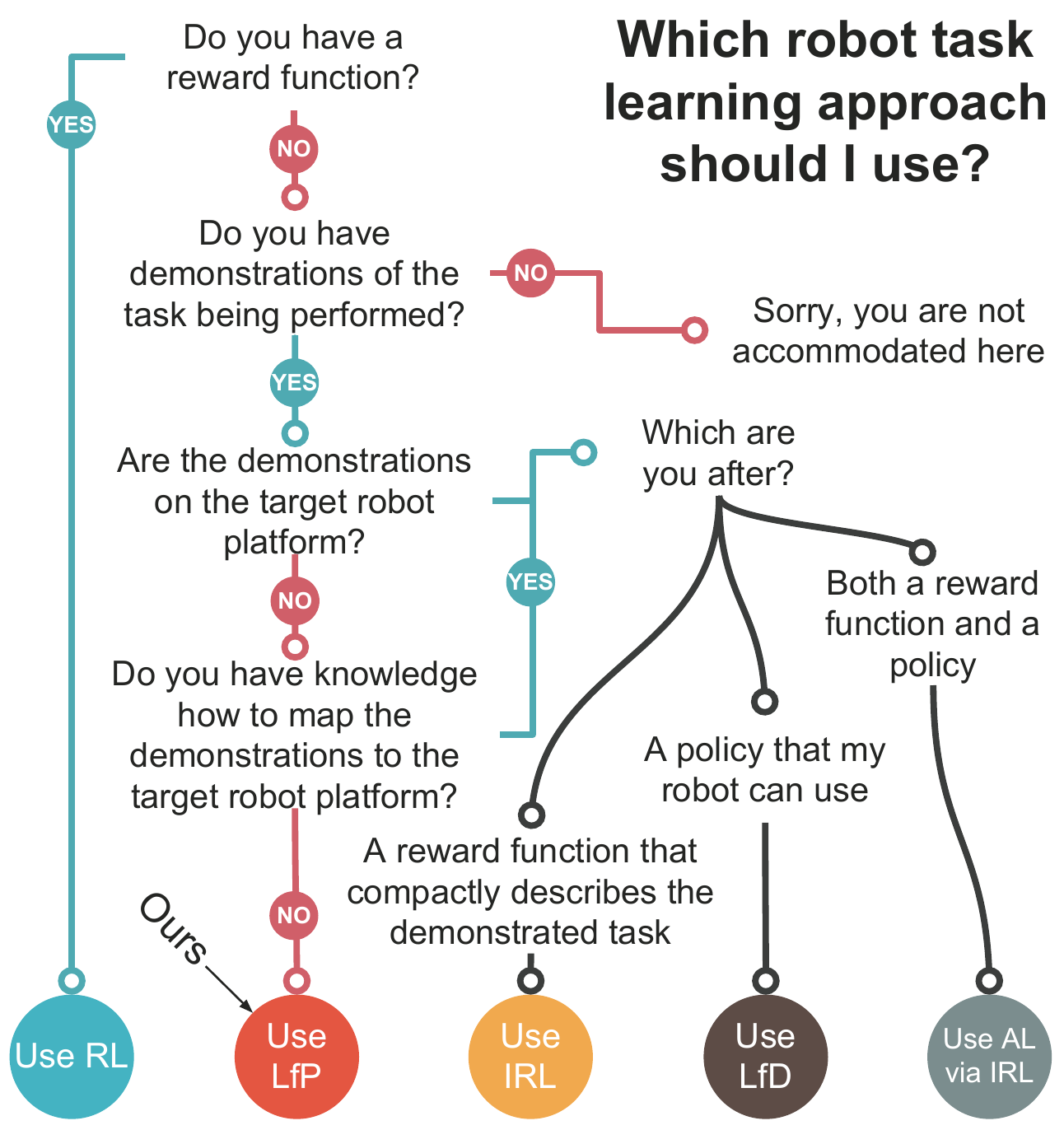}
    \caption{Choosing a robot task learning approach from Reinforcement Learning (RL) \cite{sutton1998reinforcement}, our novel Learning from Prediction (LfP) proposed in this paper, Learning from Demonstration (LfD) \cite{argall2009survey}, Inverse Reinforcement Learning (IRL) \cite{ng2000algorithms,klein2012inverse} and Apprenticeship Learning via Inverse Reinforcement Learning (AL via IRL) \cite{abbeel2007application}.}
    \label{fig:robot_task_learning_guide}
\end{figure}

Existing approaches to robot task learning generally fall under three distinct areas: Learning from Demonstration (LfD), Reinforcement Learning (RL) and Inverse Reinforcement Learning (IRL). Each approach comes with a key limitation: LfD approaches require a mapping between demonstrator kinematics and robot learner kinematics \cite{argall2009survey}, RL approaches require access to an oracle that provides rewards to the robot learner \cite{sutton1998reinforcement}, and IRL approaches require knowledge of both the states and actions executed by the demonstrator \cite{finn2016guided, abbeel2004apprenticeship, klein2012inverse}. Due to these limitations, neither RL, IRL or LfD approaches are suited for learning a task from human visual demonstrations alone - see Figure~\ref{fig:robot_task_learning_guide}.

This work is motivated by our hypothesis that many tasks can be learned by imitating the state transitions of a demonstrator alone. We investigate the case of a human demonstrator and robot learner, where the robot is able to observe the outcomes of the human actions.
Robots that can learn from human demonstrations are unquestionably a desire of many roboticists. However, to be useful in real world settings, some specific traits of any such approach are required. Firstly, the robot should generalise human demonstration sequences to unseen states; i.e.\ predict the outcome of a humans actions in states not visited during the demonstrations. Secondly, human demonstrations should be robot-agnostic; i.e.\ no knowledge or access to the target robot is required to record task demonstrations. Thirdly, the approach should be task-agnostic; i.e.\ the robot can learn new tasks provided new demonstrations alone.

We present a novel approach, termed Learning from Prediction (LfP), that addresses these requirements. Specifically, we propose to learn tasks directly from visual demonstrations by learning to predict the \textit{outcomes} of a humans actions on an environment. Operating on visual demonstrations allows for a wide range of avenues for obtaining human task demonstrations and quite naturally leads to the setting of \textbf{state} as an RGB image and \textbf{task} as sequences of RGB images. 

Our approach equips a robot with two key capabilities that enable it to imitate a task. 
Firstly, provided a small number of human-performed demonstrations, the robot can predict how the environment would look \textit{if} the human had acted in it. Secondly, provided a small number of robot-performed demonstrations, the robot can predict how the environment would look \textit{if} it acted in it. With these capabilities, the robot at each state can exhaustively search which of its actions will bring it closest to its prediction of the humans next state. 

In theory, perfect predictions will lead the robot to imitate the human perfectly across all possible states of an environment. We describe how to equip a robot with the aforementioned predictive capabilities. Our approach hinges on recent advances from the computer vision community on the problem of next-frame prediction in video sequences \cite{lotter2016deep, goroshin2015learning, mathieu2015deep, o2014learning, palm2012prediction, patraucean2015spatio, softky1996unsupervised, srivastava2015unsupervised}. Currently, next-frame predictors receive a sequence of images to predict the next image. To be useful for performing a task, we require that predictions can be made from a single image i.e. to get the robot moving from the starting location. We show that the PredNet \cite{lotter2016deep} next-frame predictor can be trained to operate on single images, if the sequences they are trained on are deterministic; i.e.\ every state has a unique next state.




We show the feasibility of our novel LfP approach on two table-top single-object manipulation tasks, designed specifically to elucidate the desirable properties of our approach. In particular, human-executed demonstrations, generality to unseen states, flexibility to different tasks, limited setup requirements and successful human-to-robot task transfer.

\section{Related Work}

\subsection{Learning from Demonstration}

The objective of Learning from Demonstration (LfD) is to teach a robot a task by demonstrating that task being performed. Specifically, the objective of LfD is to obtain a policy from example state-action pairs \cite{argall2009survey}. LfD works can be categorised by the demonstration approach taken, i.e.\ kinesthetic demonstrations, tele-operation demonstrations, or motion capture demonstrations \cite{argall2009survey, ekvall2007robot, koenig2010robot, lee2011incremental, field2016learning, schulman2016learning}. Within LfD research, solutions to the LfD problem are generally robot-specific as a result of either the demonstration approach or assumptions to map human demonstrations to the platform \cite{argall2009survey, hiratsuka2016trajectory}.

In \cite{hiratsuka2016trajectory}, a mapping between human-demonstrated movement and a robot was learned. Using a Kinect sensor and a pre-defined human model, the team showed a robot that could reproduce the demonstrated motion. While the correspondence between human and robot motions was learned, the approach required both the human model and robot to have the same dimensionality. 


While not explicitly LfD, \cite{yang2015robot} demonstrate one example of a robotic system that learns to perform tasks from demonstrations physically performed by humans. In particular, the team show how convolutional neural network (CNN) object and human action detectors can be used to produce manipulation action plans for a robot to reproduce demonstrated cooking tasks \cite{yang2015robot}. Notably, this approach requires a common action grammar between the humans and robot.

\subsection{Reinforcement Learning}

Reinforcement Learning is a widely researched approach for solving tasks formulated as Markov Decision Processes (MDPs) \cite{sutton1998reinforcement}. In recent years Deep Reinforcement Learning has been applied to a wide range of problems, including robotics \cite{levine2016end, levine2016learning, li2017deep}. At RL's core, an agent is tasked with learning a policy by exploring its environment and maximise the reward it receives from some oracle. The two key issues of applying RL to robotics applications are the exploration time and where the reward comes from. In \cite{levine2016learning}, the issue of exploration was solved by deploying 14 identical robots for 2 months to learn the task of picking up items. The issue of providing rewards 
was solved by choosing a known table height to close the gripper at and threshold the distance the gripper closes to produce reward \cite{levine2016learning}. These solutions are task-specific and engineered, disallowing the lay-person from teaching a robot a new task.

\subsection{Inverse Reinforcement Learning}

One might posit: if we can learn a policy provided a reward function, can we learn a reward function provided a policy? This idea is commonly termed Inverse Reinforcement Learning (IRL) or Inverse Optimal Control (IOC) \cite{finn2016guided}. Here the objective is to learn the underlying reward function that a demonstrator is optimising \cite{ng2000algorithms}. While it is said that a reward function is the most compact description of a task, the objective in robotics is often to find a policy such that our robot can perform the demonstrated task. 
Approaches to IRL that learn both a reward function and a policy are referred to as Apprenticeship Learning via Inverse Reinforcement Learning (AL via IRL) \cite{abbeel2004apprenticeship}. 

AL via IRL has seen use in robotics, notably for aerobatic helicopter flight \cite{abbeel2007application}. In this case, recordings of an expert remotely piloting a helicopter were used to learn the weights for a hand-crafted 24-feature vector that defined the reward function for the task. More recently, \cite{finn2016guided} applied IOC to a number of real world robotic manipulation tasks. In this case, a neural network was used to express a reward function, removing the need for hand-engineering reward features \cite{finn2016guided}. Demonstrations of the task being performed were provided by kinesthetic teaching and as such, these demonstrations are tied to the robot platform they were performed on.

\subsection{Video Prediction using Deep Networks}

Next-frame video prediction is an unsupervised learning problem studied in the computer vision community \cite{lotter2016deep, goroshin2015learning, mathieu2015deep, o2014learning, palm2012prediction, patraucean2015spatio, softky1996unsupervised, srivastava2015unsupervised}. The objective from a computer vision perspective is to leverage the wide-spread availability of video to learn feature representations that are useful for solving other tasks, i.e. object detection. The PredNet architecture used herein is one such algorithm designed for next-frame video prediction \cite{lotter2016deep}. PredNet is comprised of a number of stacked modules that attempt to predict the input to that module. PredNet is shown to perform well on both synthetic and real world tasks and can support variable length inputs at test time due to internal recurrent layers. 

Video prediction techniques have also been applied to help solve Reinforcement Learning problems. \cite{oh2015action} presented a deep architecture that could learn to predict frames in Atari 2600 games. By using the actions of the player within the network, the team were able to predict the state of the game up to 100 frames into the future. The team did not use the predictions to improve game-play performance however demonstrated that the game could be played on their predictions alone \cite{oh2015action}.

More recently, a deep architecture was presented that won the Full Deathmatch track of the Visual Doom AI Competition \cite{dosovitskiy2016learning}. In their work, in-game measurements such as health, ammunition and score are combined with the current image and the current goal to inform action selection. The team use action-conditioned predictions of the in-game measurements to select the action that it predicts will bring it closest to the current goal. Note it is assumed that the goal can be represented as a function of these predicted measurements.

In the case of robotics, a robotic pushing dataset was presented in \cite{finn2016unsupervised} alongside a new approach for predicting the appearance of the environment conditioned on a robot's actions. 

\subsection{Summary}

The existing areas of LfD, RL and IRL have yet to address the problem of robots learning from demonstrations where a human performs the task and no mapping between the human and the robot is made. Applications of video prediction techniques to RL and robotics have focused on improving next-frame predictions by conditioning predictions on the agent or robots actions. An existing approach that used prediction techniques for action selection relied on additional information over raw images, with training and testing performed by the same agent \cite{dosovitskiy2016learning}.

We propose to learn tasks directly from visual demonstrations by learning to predict the outcome of human and robot actions on an environment, without access to the thought processes or actions of the human.

\section{Learning from Prediction}

We herein present the Learning from Prediction approach. For generality, we define our human demonstrator as \textbf{Expert} and our robot as \textbf{Agent} in this section.

\subsection{The Expert}
Let us assume a deterministic function $\vpi^E:\mathbb{S} \rightarrow \mathbb{U}$ describes the actions $\vu_t \in \mathbb{U}$ chosen by an expert when in state $\vs_t \in \mathbb{S}$, so that $\vu_t = \vpi^E(\vs_t)$. This is typically called a \emph{policy}. Likewise, a probabilistic model $p(\vs_{t+1} | \vs_t, \vu_t)$ describes the transition from one state into the next, given action $\vu_t$ was executed.

The thought processes leading a human expert to choose action $\vu_t$ are unobservable to a robotic agent. In fact, we argue that even the expert's action space $\mathbb{U}$ is unknown and inaccessible to the agent. This results in both  $\vpi^E$ (the expert's internal decision process) and $p(\vs_{t+1} | \vs_t, \vu_t)$ (a model of how the world reacts to the expert's actions) being inaccessible. 

However, the agent can observe the state of the world while the expert is acting under its policy $\vpi^E$. The occurring state transitions $\vs_t \rightarrow \vs_{t+1}$ are observable, assuming the robot is equipped with the appropriate sensors. These state transitions under the policy are described by the distribution $p(\vs_{t+1} | \vs_t, \vpi^E)$. 

While a full probabilistic model of the true distribution $p(\vs_{t+1} | \vs_t, \vpi^E)$ is hard to learn, we demonstrate that approximating a deterministic predictive model $P : \mathbb{S} \rightarrow \mathbb{S}$, so that $P(\vs_t) =  \argmax_{\vs_{t+1}} p(\vs_{t+1} | \vs_t, \vpi^E)$, is computationally tractable.

\subsection{The Agent} \label{theory_the_agent}
Our goal is to train an agent to choose actions $\va_t \in \mathbb{A}$ according to a parametric policy $\pi_\vtheta(\vs_t)$. That is, we seek the optimal model parameters $\vtheta^*$ based on an optimality criterion yet to be defined.

Notice that the agent's actions $\va_t$ are elements of the action space $\mathbb{A}$, while the expert's actions $\vu_t$ are elements of $\mathbb{U}$. The two spaces $\mathbb{A}$ and $\mathbb{U}$ are not identical. This makes sense since the actions that can be performed by a human will often differ greatly from the action space available to a robot\footnote{Notice that this concept extends naturally to the case where the expert is another robot or technical system, with an action space incompatible to that of the agent that is to be trained.}.

Similar to above, a model $q(\vs_{t+1} | \vs_t, \va_t)$ describes how the state of the world changes when the agent is acting in it. As discussed before, obtaining the full probabilistic model is intractable, but we can utilise PredNet to approximate a predictive model $Q : \mathbb{S} \times \mathbb{A} \rightarrow \mathbb{S}$, so that $Q(\vs_t, \va_t) =  \argmax_{\vs_{t+1}} q(\vs_{t+1} | \vs_t, \va_t)$.

\subsection{Finding the Optimal Policy}
Since we assume the expert chooses optimal actions, we would ideally like to mimic the expert's behaviour. However, since the action spaces $\mathbb{A}$ and $\mathbb{U}$ are incompatible, and the expert's actions $\vu_t$ are unobservable as discussed above, it is impossible to learn a direct mapping from $\vu_t$ to $\va_t$. 

Instead, we propose the following optimal policy:
\begin{equation}
    \vpi^*(\vs_t) = \argmin_{\va_t^{(i)}} \tilde P(\vs_t) \ominus \tilde Q(\vs_t, \va^{(i)})
\end{equation}
This policy executes the optimal action $\va^*_t$ that minimises the difference between the \emph{predicted outcome} of the expert acting in state $\vs_t$, and the \emph{predicted outcome} of the agent executing $\va_t$ in the current state. We write $\ominus$ above to indicate a suitable difference metric on the state space $\mathbb{S}$.

In this paper we utilise PredNet \cite{lotter2016deep} to learn the approximations $\tilde P(\vs_t)$ and $\tilde Q(\vs_t, \va^{(i)})$ and train it directly on raw images. The state space $\mathbb{S}$ therefore is the space of RGB images, and we show that a suitable metric to implement the $\ominus$ operator is the mean squared error between the raw pixel values. We choose to operate on raw images to maintain the robot-agnostic and task-agnostic traits of our approach.


\section{Experimental Setup} \label{experimental_setup}

We wish to train a robot to execute a task by imitating the state transitions of human demonstrations. The humans internal decision process and a model of how the world reacts to the humans actions is unknown. We demonstrate an approach for approximating the model $p(\vs_{t+1} | \vs_t, \vpi^E)$ provided only images of human-executed task demonstrations. We also demonstrate that this approach can:

\begin{itemize}
  \item  generalise observed demonstrations to predict how the environment would change if the demonstrator had acted in an unseen state
  \item be applied to different tasks without modification to the approach
\end{itemize}

We restrict our investigation to two table-top, single-object manipulation tasks that demonstrate the desirable traits of our approach. The first task requires the target object to be moved to a target location and is referred to as \textit{movetopos}. The second task requires the target object to be moved in a specific direction based on its spatial location and is referred to as \textit{pushpull}. These tasks can be considered building blocks of more complicated tasks that involve multiple objects, such as \textit{clean the table}.




\subsection{The Environment}

\begin{figure}[t!b]
\centering
  \subfloat[Discretised Task Space]{\label{fig:grid_state_space}
    \includegraphics[width=0.68\linewidth]{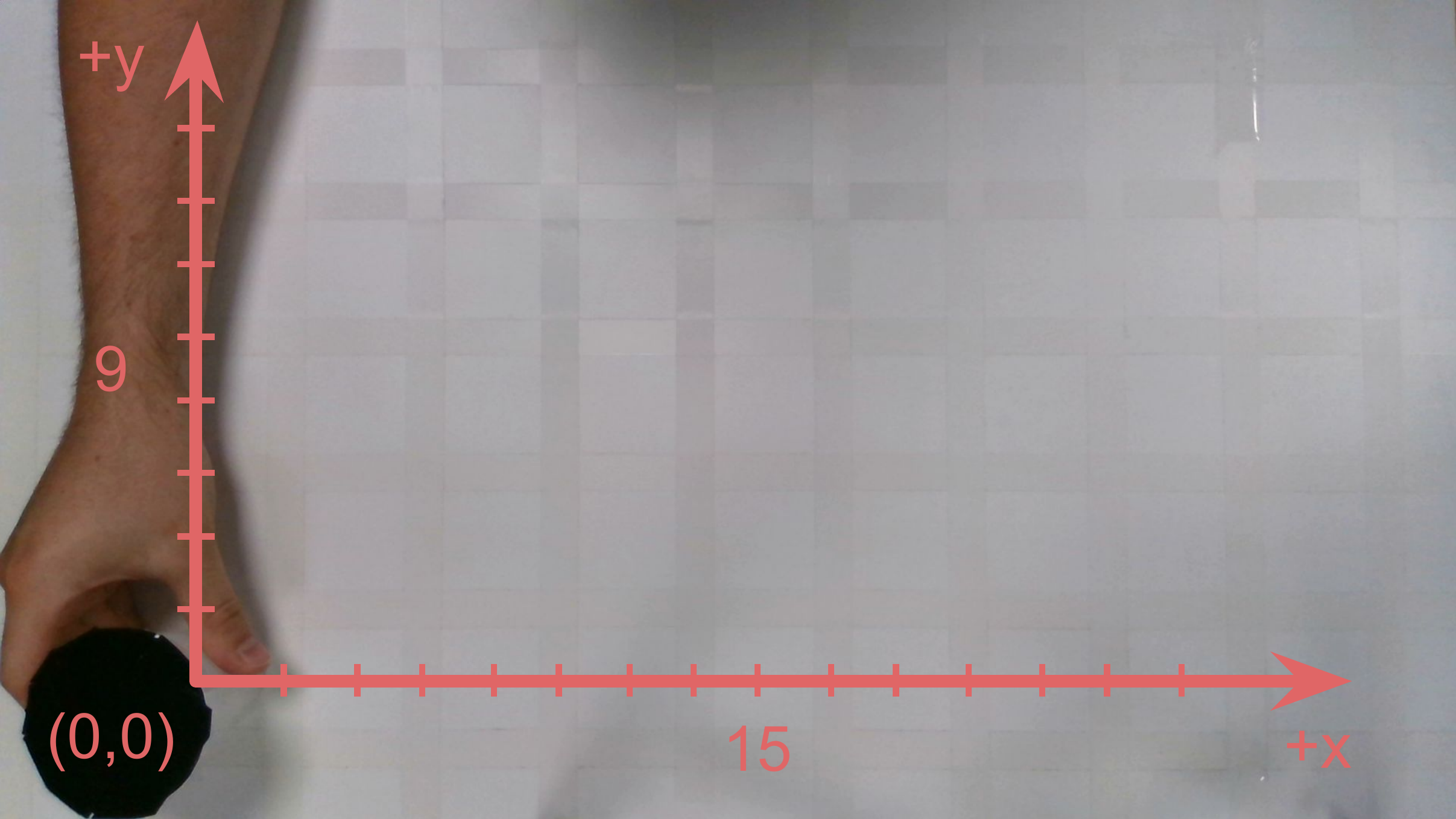}}\hfill
  \subfloat[Robot and Object]{\label{fig:robot_holding_object}
    \includegraphics[width=0.2685\linewidth]{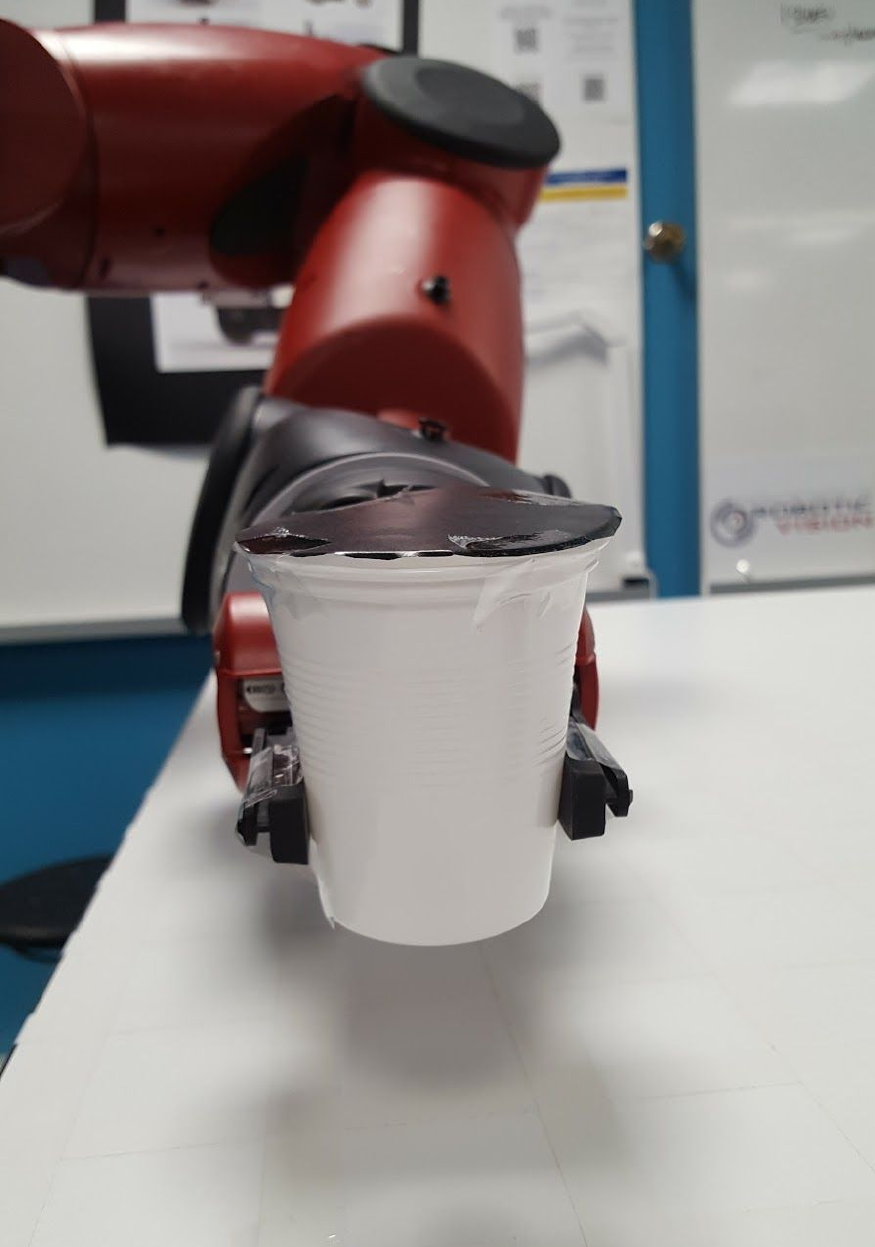}}\\
  \caption{(a) Human holds the object at the origin of the discretised (15x9 grid) task space. The Object can be positioned anywhere within. A tripod-mounted RGB camera provides the overhead view. (b) Learner agent is a Baxter robot, pictured holding the round, black object above the table-top environment.}
\label{fig:environment_setup}
\end{figure}

We consider the situation where a human would like to teach a robot, or robots, a set of single-object manipulation tasks. Specifically, we have a table-top environment (that is discretised into a 15x9 grid) and a round, black object that can be moved around the space (Figure~\ref{fig:environment_setup}). An overhead RGB camera is used to record task demonstrations. Demonstrations are composed of image sequences that are synchronised with state transitions through the grid. Repeatability of demonstrations and starting locations is facilitated by a white grid structure that is partially visible in the figures. We assume the object position to be discrete.

We consider only generalisation to states unseen in the demonstrations. As such, we assume that the discretised task-space, lighting conditions, object, camera and camera location remain unchanged for both human and robot interaction.

\subsection{The Expert}

We wish to approximate human task demonstrations with PredNet and demonstrate generalisation to unseen states. As mentioned earlier, we have selected two task variants based on our common grid environment to demonstrate our robot task learning approach. 

The first task variant is known as \textit{movetopos}: it demonstrates an example where an object is moved from an arbitrary start location to a desired final location.  
Successfully reaching the goal location from unseen start locations indicates generalisation. 

The second task variant is known as \textit{pushpull}: it demonstrates an example where a different action must be performed based on the spatial location of the object. In our scenario, if the object is positioned in the upper half of the grid space, we wish it to move right, if in the lower half, move left.
Moving in the correct direction when starting in rows of the grid unseen during the demonstrations indicates successful generalisation.

\begin{figure}[t!b]
\centering
  \subfloat[\textit{pushpull}]{\label{fig:pushpull}
    \includegraphics[width=0.48\linewidth]{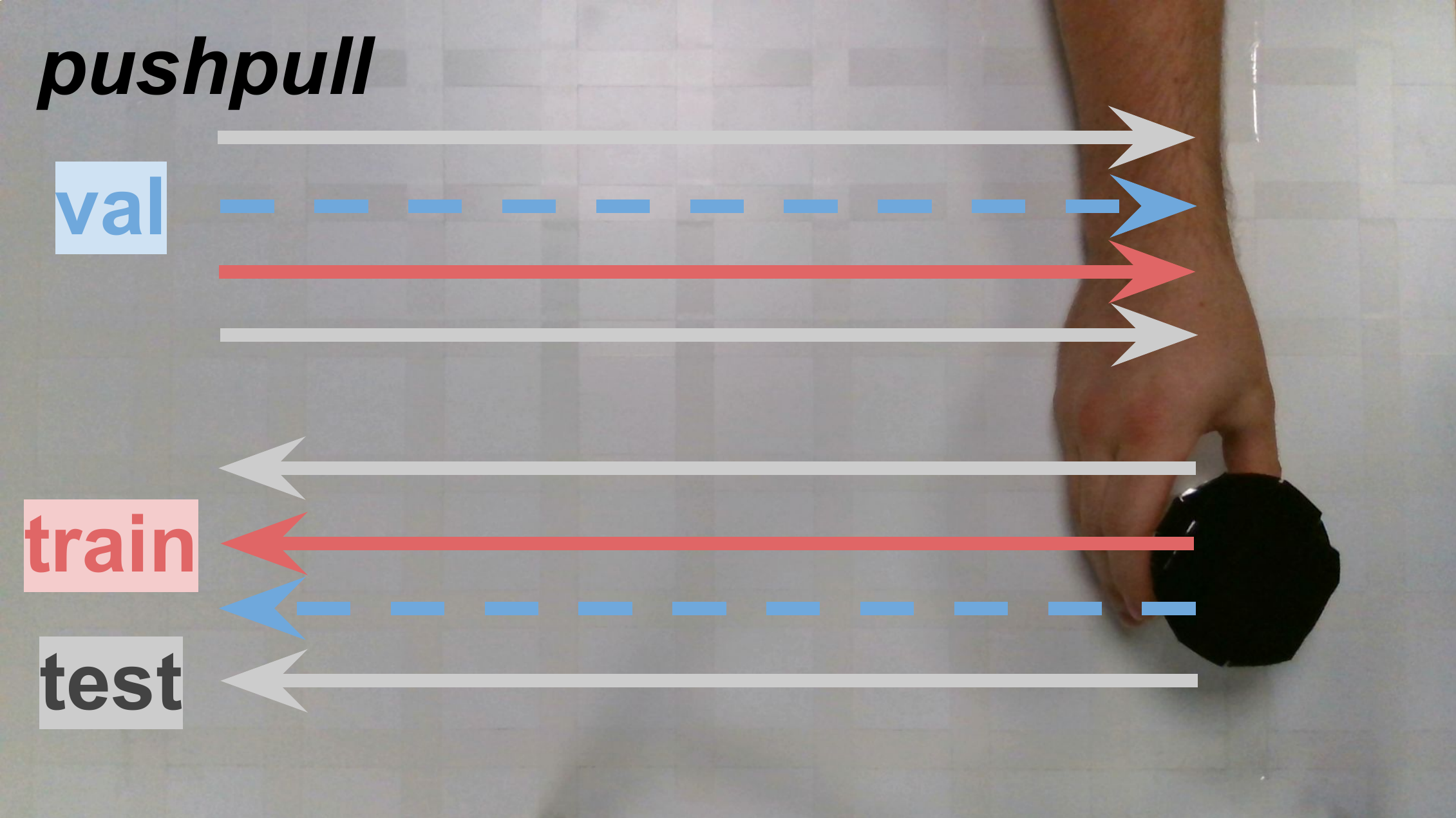}}\hfill
  \subfloat[\textit{movetopos}]{\label{fig:movetopos}
    \includegraphics[width=0.48\linewidth]{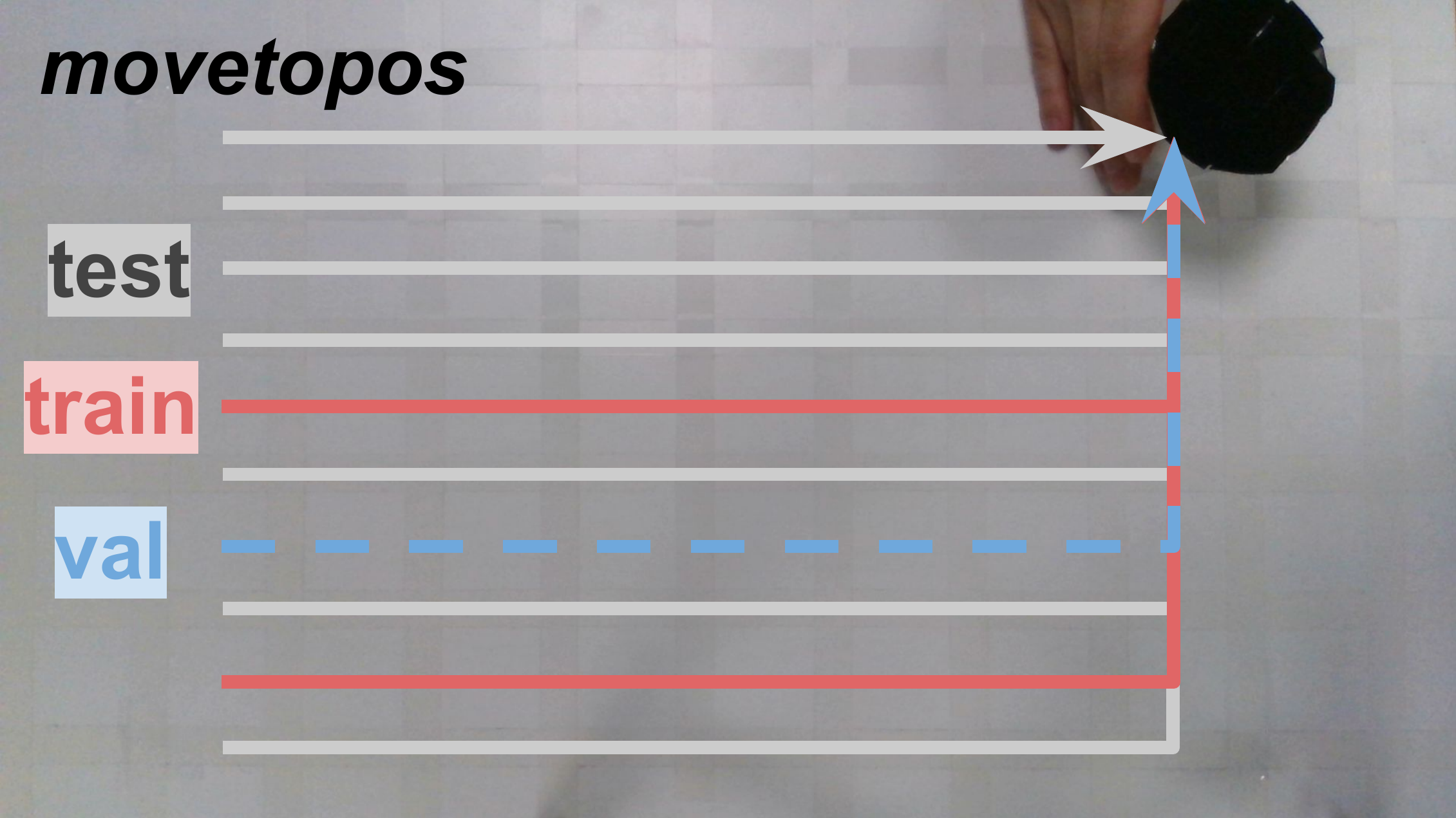}}\\
  \caption{(a) The \textit{pushpull} task involves moving the object left or right based on its spatial location. (b) The \textit{movetopos} task involves moving the object to a specific location; the upper right-hand corner in our case.}
\label{fig:tasks}
\end{figure}

We use PredNet to approximate the predictive model $P(s_t)$ of the demonstrator. Recall that $P(s_t)$ is a prediction of the demonstrators next state provided the current state. We train PredNet to predict our human demonstrators actions by providing sequences of images that capture the desired task being performed. To collect sequences of images, we hold the object as a human and move the object through the grid space in discrete steps. An image of the environment is captured after each move is performed. Notice that the human's arm remains in the view of the scene. The specific training and validation sequences recorded for each of the two tasks can be seen in Figure~\ref{fig:tasks}.

We trained one PredNet for each task and primarily maintained the hyper-parameters reported by its authors on the Kitti dataset \cite{lotter2016deep}. 
While we use a separate PredNet for each action herein, prior approaches that produce action-conditional predictions with a single network may alternatively be used \cite{oh2015action, dosovitskiy2016learning, finn2016unsupervised}. It is unclear how data-efficient action-conditional networks can be as reported applications involve training on thousands of frames. In our current setting, using multiple networks is not a limitation. We found a sequence length of five frames resulted in successfully capturing the change in direction within the \textit{movetopos} task. We used a batch size of 4 image sequences and processed 64 image sequences per epoch (samples per epoch = 64). We trained the network for a maximum of 500 epochs, only keeping the network weights that performed best on the validation sequences. While not reported, we briefly trialled different sequence lengths and noticed the predictions no longer captured the direction change correctly, highlighting the need for a separate investigation into hyper-parameter selection and potentially into different prediction frameworks.

\subsection{The Agent} \label{experimental_setup_the_agent}

We wish to approximate the results of a robot's primitive actions on the environment with PredNet and demonstrate generalisation to unseen states. Specifically, we use a separate PredNet to approximate each of the action-specific predictive models $Q(\vs_t, \va^{(i)})$ of the robot. Recall that $Q(\vs_t, \va^{(i)})$ is a prediction of the robot's next state provided the current state and taking action ‘$a$’.

As mentioned in Section \ref{theory_the_agent}, we propose that the action spaces of the expert and agent are incompatible, $\mathbb{A} \neq \mathbb{U}$. Under this assumption, we state that predictions of the expert should be made on the current state alone, $P(s_t)$. In practice, we found PredNet was incapable of producing accurate single-image predictions for all states; leading to poor overall performance at the tasks. To remedy this result, we selected the action space $\mathbb{A} = \{up, down, left, right\}$ for the robot and aligned this with the action capabilities of the human. As such, the distance the object moves under both human and robot actions was the same. Employing this setting allowed the states physically visited by the robot to be used in the prediction of the human's next state, $P(s_{0 \rightarrow t})$. 



\begin{figure}[t!b]
\centering
  \subfloat[Robot up primitive]{\label{fig:primitive_up}
    \includegraphics[width=0.48\linewidth]{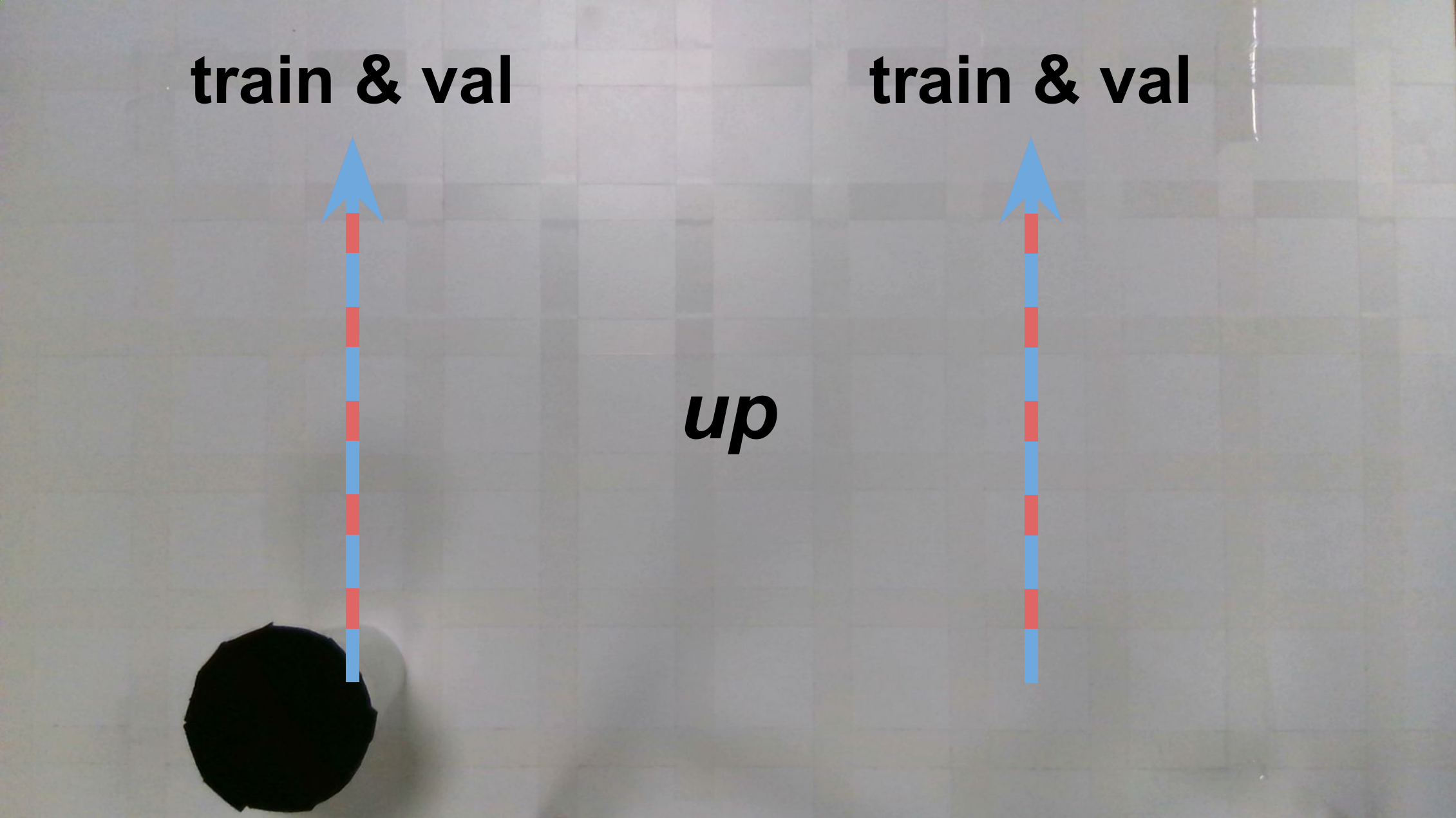}}\hfill
  \subfloat[Robot down primitive]{\label{fig:primitive_down}
    \includegraphics[width=0.48\linewidth]{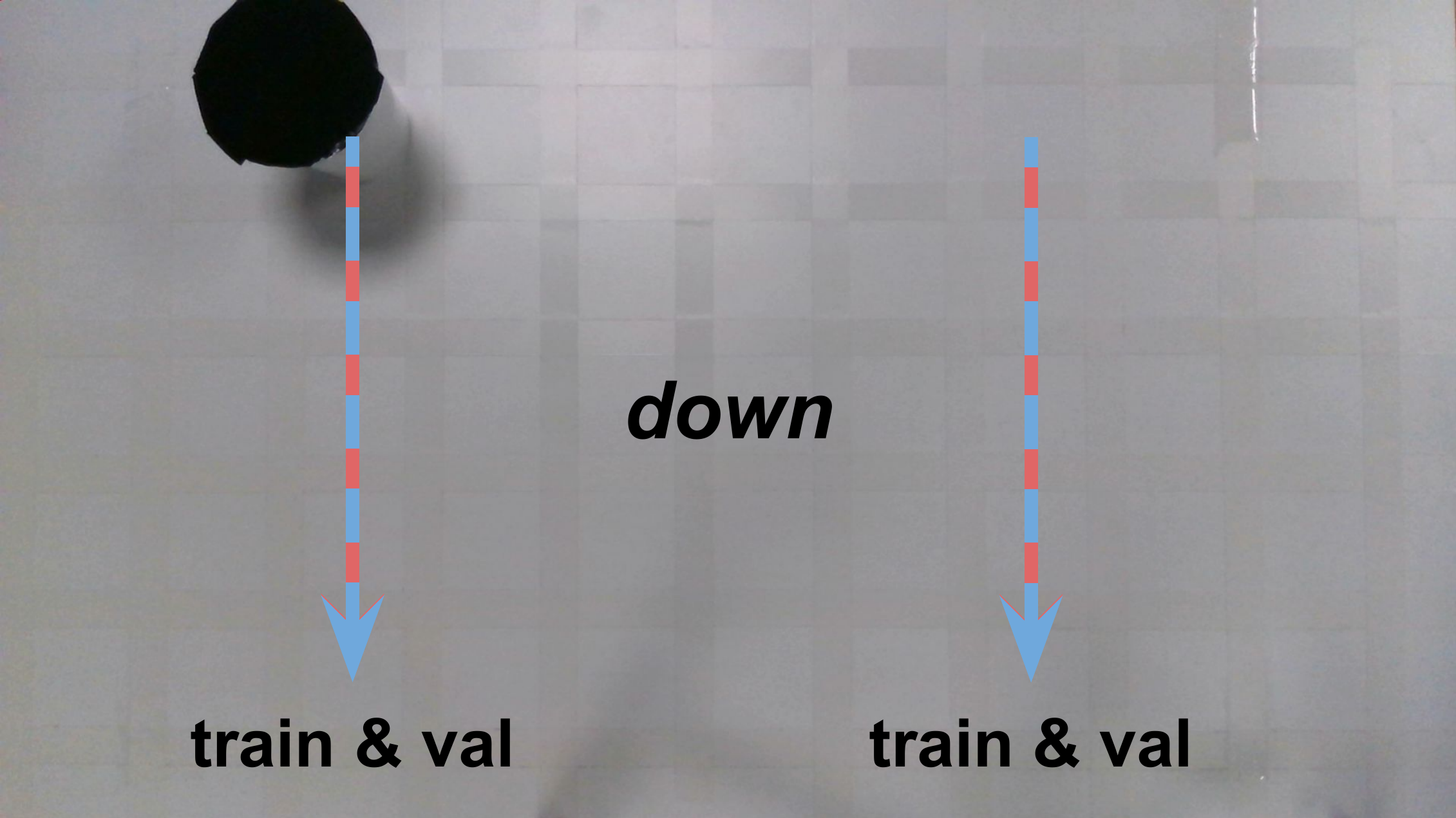}}\\
  \subfloat[Robot left primitive]{\label{fig:primitive_left}
    \includegraphics[width=0.48\linewidth]{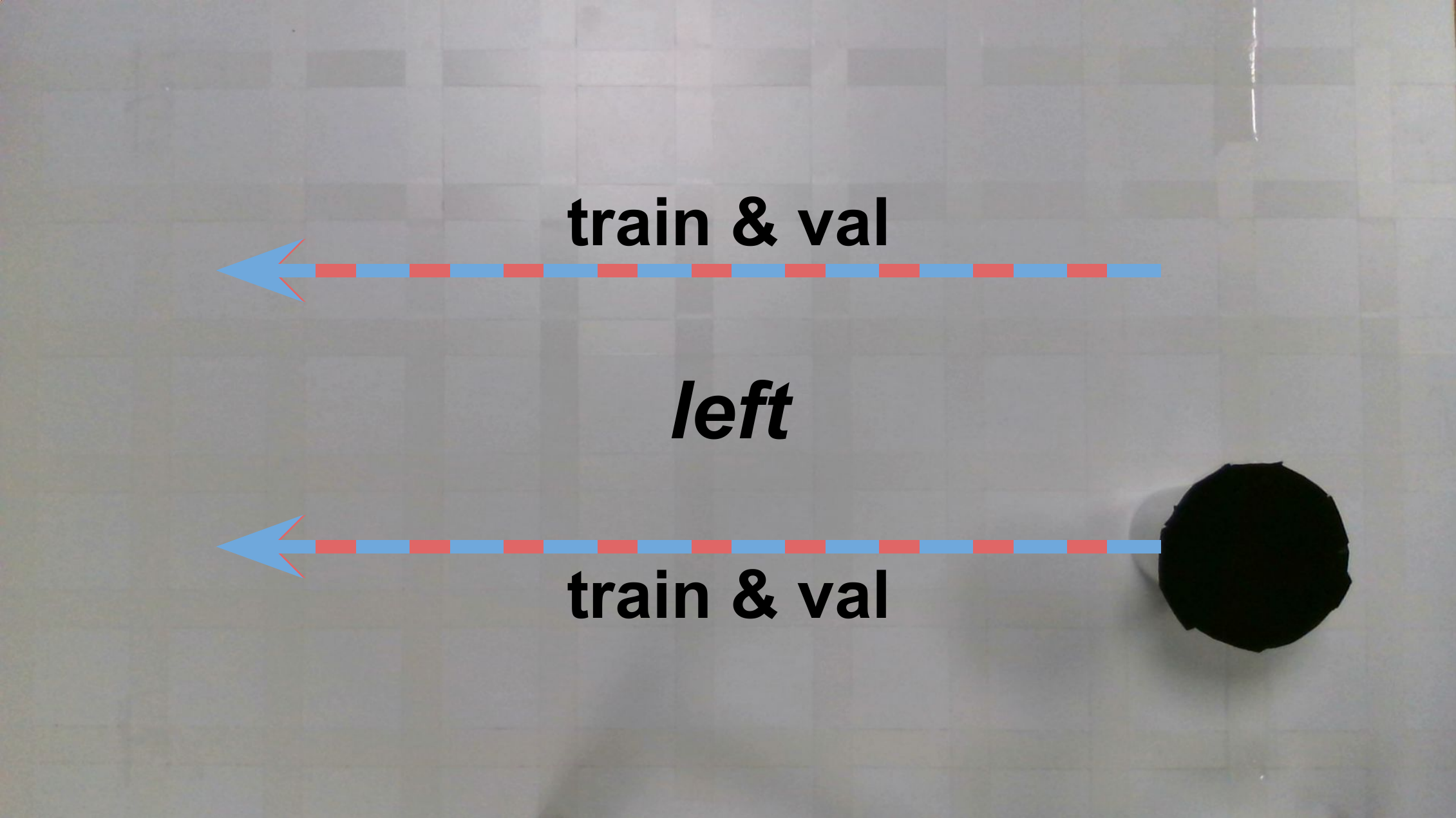}}\hfill
  \subfloat[Robot right primitive]{\label{fig:primitive_right}
    \includegraphics[width=0.48\linewidth]{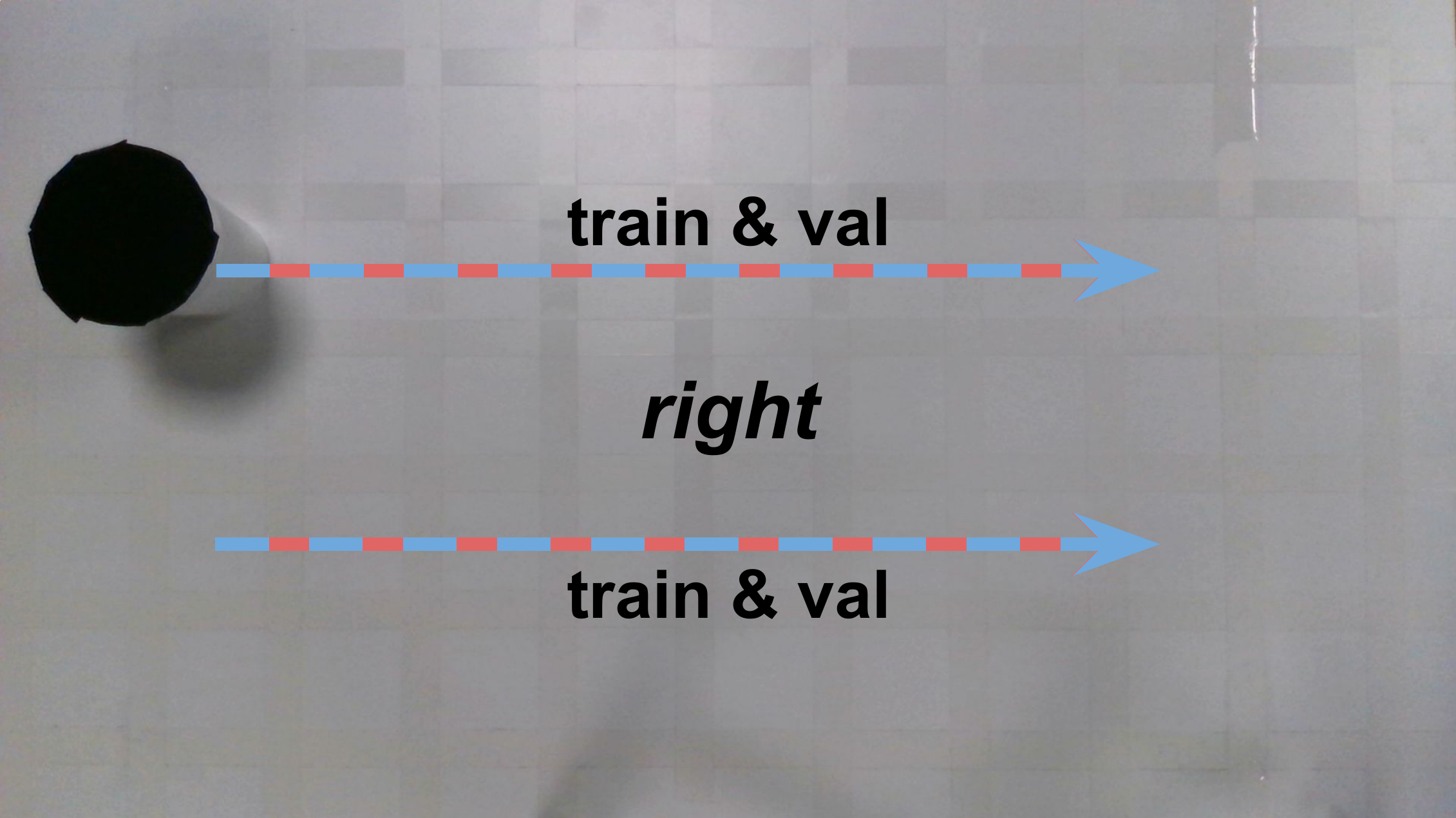}}\\
  \caption{Robot action primitives were trained using the demonstration trajectories depicted. The robot's arm is not present in the image to remove bias as to how the object should be held at test time. While the arm was removed artificially herein, the process can be automated with a calibrated camera, robot model and image of the empty scene.}
\label{fig:action_primitives}
\end{figure}

With the four primitive actions chosen, we collect training data that captures how the action primitive influences the state of the environment. The training data was collected by having the robot perform each of its action primitives twice across the grid space. The training and validation data used is depicted in Figure~\ref{fig:action_primitives}. Note for both collecting the robot primitive training data and implementing the approach on the robot, we require that the robot arm be removed from the image of the scene. By removing the robot arm, we remove any bias of the predictions prescribing the robots joint configuration while performing the task. In this work, we achieve this by taking images of the item in each grid location without the robot arm. This can trivially be replaced by a segmentation routine based on a known camera pose and robot model, coupled with a background image of the environment. We argue that this requirement is reasonable of current-day robot systems. NB: we do not require the human arm to be removed from the scene.

\subsection{Finding an Optimal Policy} \label{experimental_setup_finding_optimal_policy}

We now have four action primitive PredNet’s and two task-specific PredNet’s. Provided the current state, the action primitive PredNet’s provide a prediction of the environment as if the corresponding action had been performed by the robot. Ideally, the task-specific PredNet’s would also operate off the current state alone, and provide a prediction of the environment as if the human demonstrator had performed an action. By predicting off the current image alone, there is no requirement for the sequence of state transitions previous to the current state to align with the sequence of state transitions demonstrated by the human. As mentioned in the previous section, we boosted prediction performance by providing the robot’s physically executed state transitions into each subsequent prediction of the humans next state. 

Our approach for applying the task-specific and action primitive predictors on a real robot platform is captured in Algorithm \ref{algo:max}. The action primitive predictors allow the robot to choose the action which minimises the difference between its next state, and its prediction of the humans next state. Assuming successful predictions, our algorithm results with the robot successfully imitating the human on the first attempt of the task.

\begin{algorithm}[t!b]
\SetAlgoLined
\DontPrintSemicolon
$state\_sequence \gets []$\;

\For{$i \gets 0$ \textbf{to} $sequence\_length$}{
    $current\_state \gets capture\_image()$\;
    $state\_sequence.append(current\_state)$\;
    $P(s_t) \gets predict\_expert(state\_sequence)$\;
    
    \For{$a$ \textbf{in} $[up, down, left, right]$}{
        $Q(s_t, a) \gets predict\_action(current\_state, a)$\;
        $errors[a] \gets MSE( P(s_t), Q(s_t, a) )$\;
    }
    
    $action \gets \displaystyle\argmin_a(errors)$\;
    $perform\_action(action)$
}
\caption{Learning from Prediction.
}
\label{algo:max}
\end{algorithm}

\section{Results}




\begin{table*}[t!b]
\centering
\caption{Performance at each task from all possible starting locations.}
\label{results_table}
\begin{tabular}{lll}
\toprule
~                                                                       & pushpull                 & movetopos               \\
\midrule
Successful trajectories (predictions use previous sequence)                                 &  74.1\%                  & 100.0\%                 \\
Successful trajectories (predictions from single-image)								&  37.5\%                  &  10.4\%                 \\
\midrule
Successful trajectories with no deviation from ground truth       &  96.4\%                  &  62.7\%                 \\
Length of deviations from ground truth (additional steps taken to reach goal)       & median 6, max 28, min 5  & median 1, max 1, min 1  \\
\bottomrule
\end{tabular}
\end{table*}

We report the overall performance of Learning from Prediction on our two proposed tasks in Table \ref{results_table}. For each task, we tested the system from every possible start location, excluding their goal locations. We define a trajectory as a sequence of steps the robot is allowed to move the object within the task environment. A trajectory is successful if the object arrives at the ground truth final location, in alignment with the demonstrations, see Figure \ref{fig:tasks}.

100\% of the 135 trajectories for the \textit{movetopos} task successfully arrived at the goal location in the top right-hand corner of the task-space. 74.1\% of the 112 trajectories for the \textit{pushpull} task successfully arrived at the goal locations. In addition to this primary result, we report the performance of the approach using single-image predictions alone. We find significantly lower performance in the single-image case. Recall from Section \ref{experimental_setup_the_agent} that to improve over the single-image trajectory performance, we fed all previously visited states into each subsequent next state prediction.


Secondary to the percentage of successful trajectories, we report the percentage of trajectories that deviated from the ground truth. Each starting location has a fixed path to the goal as demonstrated by the human. The \textit{movetopos} task had a significant number of trajectories with an additional step. In these cases, poor predictions at the transition from moving rightwards to upwards delayed the moving up by one step. Feeding the sequence into the predictor at this failure location lead to the significant performance increase for the \textit{movetopos} task against the single-image trajectories. 


\subsection{Action-Primitive Prediction Performance}

\begin{figure}[t]
    \centering
    \includegraphics[width=\linewidth]{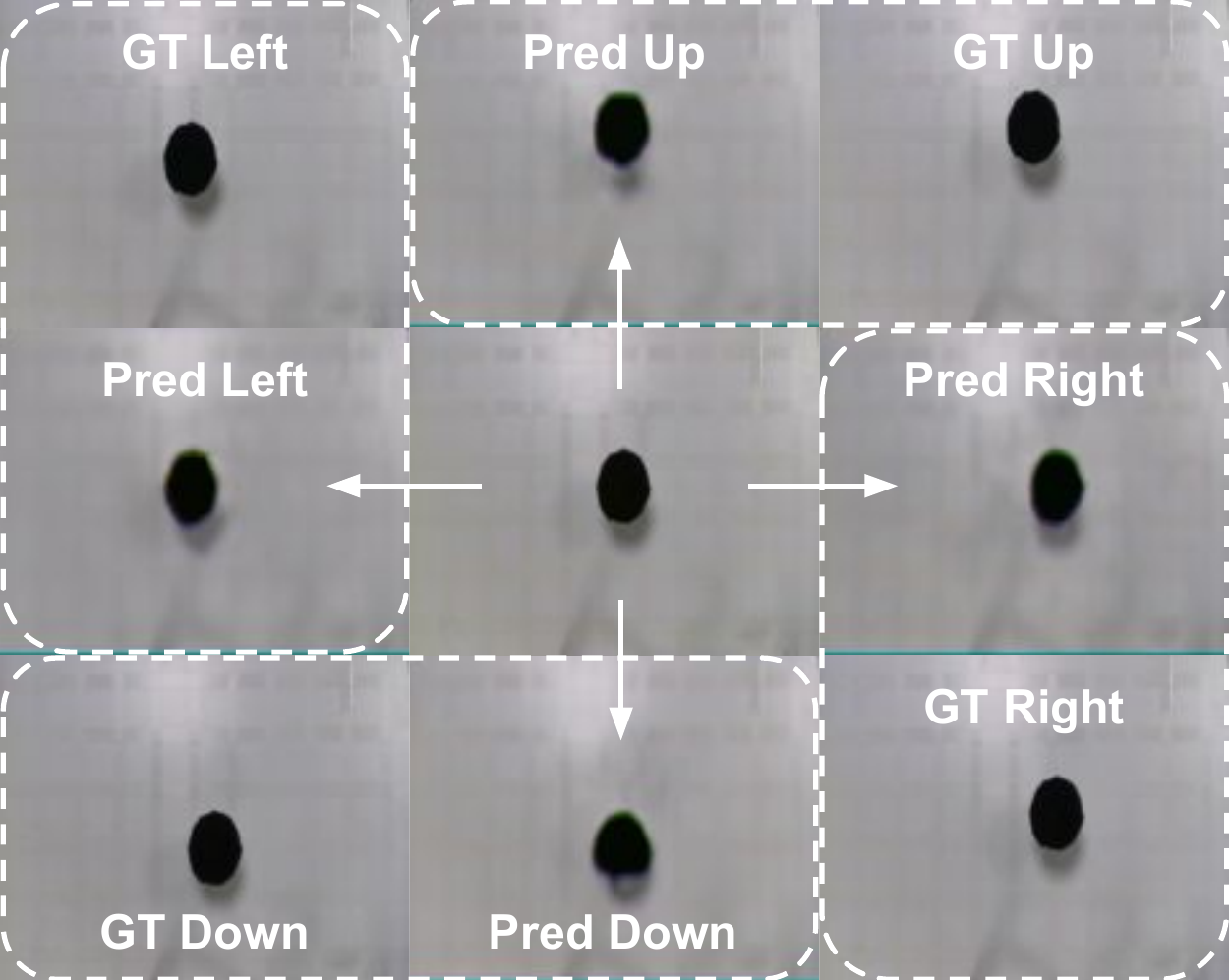}
    \caption{Examples of action primitive predictions vs. corresponding ground truth. The four action primitive PredNets were trained as per Figure \ref{fig:action_primitives}. 
}
    \label{fig:action_primitives_prediction_results}
\end{figure}

Action-primitive prediction allows the robot to predict how the environment would look \textit{if} it acted with a given action. We trained four action-primitive predictors as per Section \ref{experimental_setup_the_agent}. We show prediction performance compared against ground truth for an unseen part of the task-space in Figure \ref{fig:action_primitives_prediction_results}. As can be seen, the predictions align very well with the ground truth. Only two trajectories across the task space were required to train each action primitive network - see Figure \ref{fig:action_primitives}.

\subsection{Task Performance}

\begin{figure}[t!b]
\centering
  \subfloat[\textit{pushpull}]{\label{fig:pushpull}
    \includegraphics[width=0.98\linewidth]{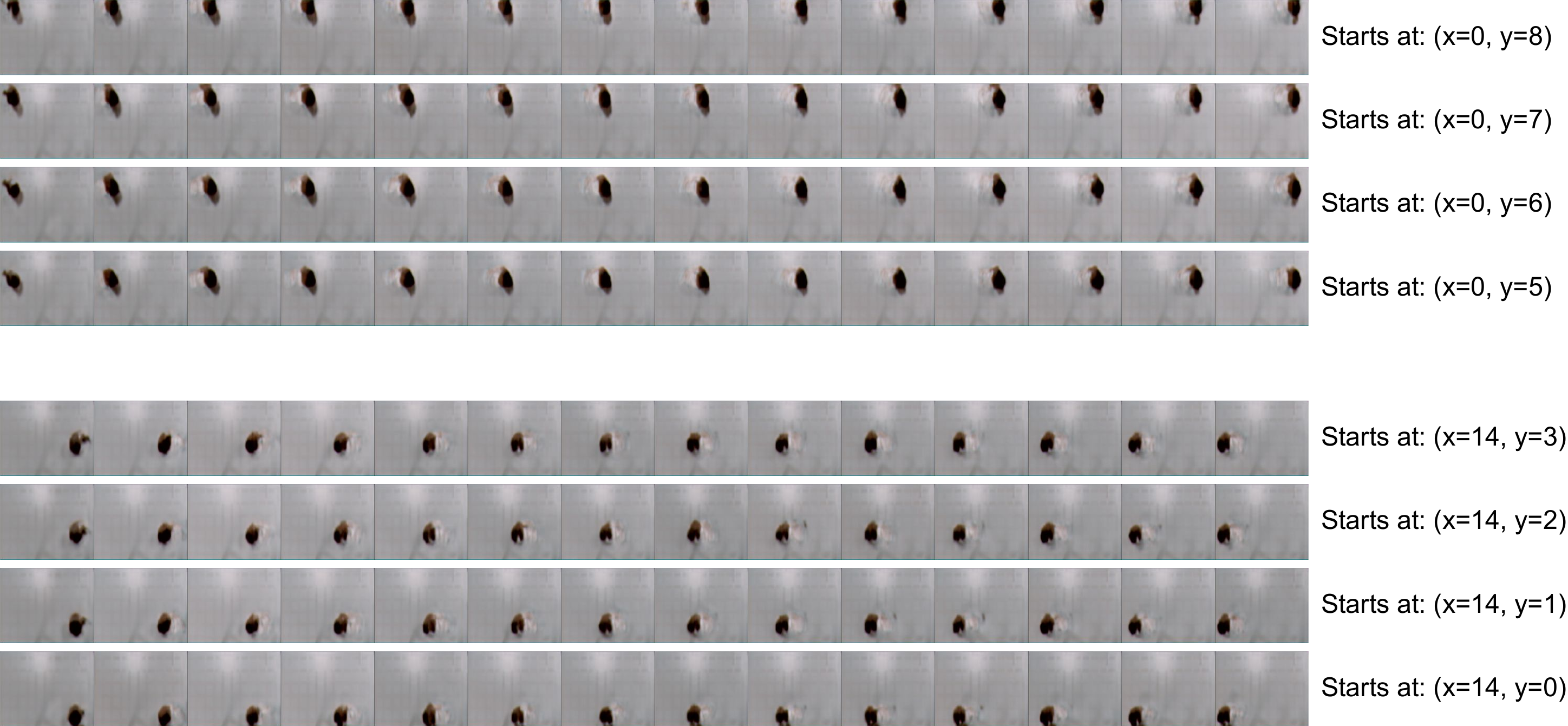}}\\
  \subfloat[\textit{movetopos}]{\label{fig:movetopos}
    \includegraphics[width=0.98\linewidth]{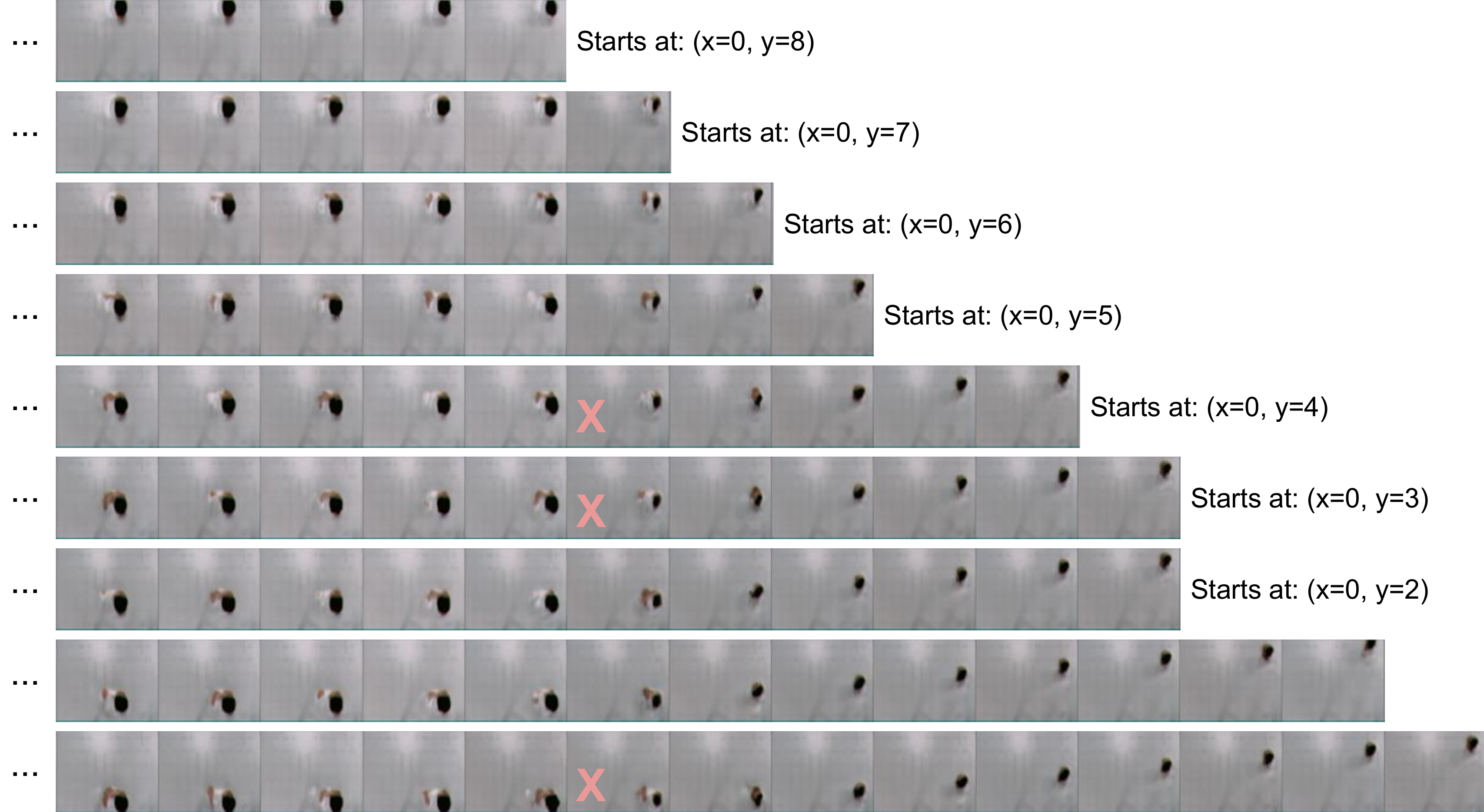}}\\
  \caption{Full sequences of predicted images as exemplars are shown. The red x's in the \textit{movetopos} sequences mark predicted images that resulted in an incorrect action selection. This caused these sequences to take one additional step over the ground truth. Overall, 50 of the 134 successful \textit{movetopos} trajectories contained an additional step at the transition point from moving rightwards to upwards. The MSE between action primitive prediction right and up at the failure locations were very close.}
\label{fig:task_results}
\end{figure}
 
Task prediction allows the robot to predict how the environment would look \textit{if} the human acted. We show a number of full sequences of predicted images as exemplars in Figure \ref{fig:task_results}. As can be seen, the predictions move the block across the task-space in alignment with the human-performed demonstrations. Note that a number of the states visited in these exemplars were not visited by the human.

\subsection{Failed Trajectories and First-State Prediction Performance}

\begin{figure}[t]
    \centering
    \includegraphics[width=\linewidth]{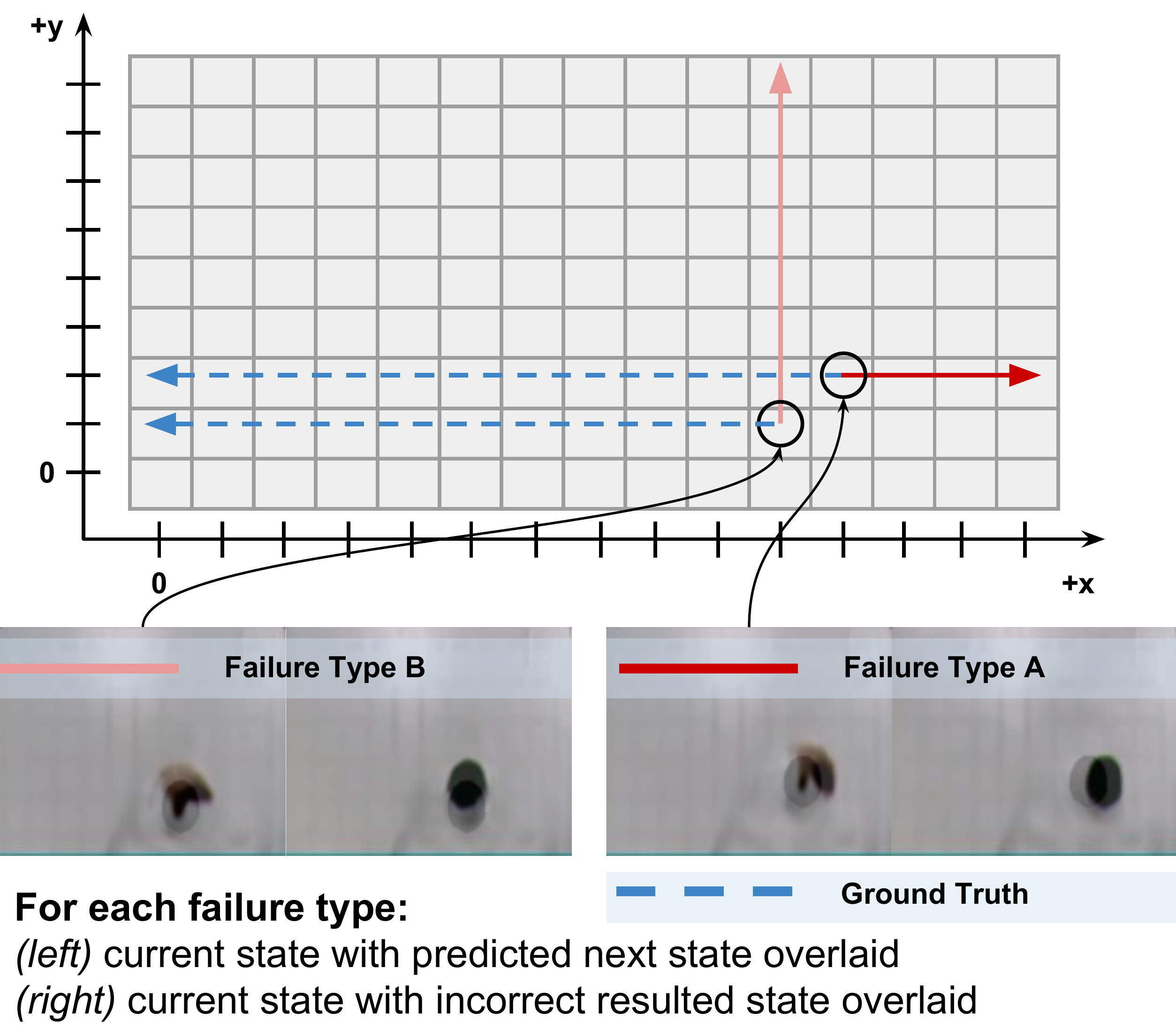}
    \caption{The \textit{pushpull} task had 29 unsuccessful trajectories of a possible 112. 26 of the failures (Type A) incorrectly moved the item right on the first prediction. 3 of the failures (Type B) moved the item upwards. These two exemplars highlight the poor first-state predictions that caused these failures. Recall for the \textit{pushpull} task the objective is to move the item left if it is located in the lower half of the task-space.}
    \label{fig:major_failure_exemplar}
\end{figure}

25.9\% of the \textit{pushpull} trajectories failed. Of these trajectories, we found that deviation from ground truth occurred at the first state with an incorrect prediction as per Table \ref{results_table}. 26 of the trajectories failed by moving the object rightwards from the starting state (as opposed to the ground truth leftwards). 3 of the trajectories failed by moving the object upwards from the starting state. Visual observation of the single-image predictions at these starting locations highlights why the incorrect action was taken - see Figure \ref{fig:major_failure_exemplar}.

\begin{table*}[t!b]
\centering
\caption{We report single-image prediction performance irrespective of the final trajectories success. We report how making a correct prediction in the first state relates to the success of that trajectory.}
\label{results_table_2}
\begin{tabular}{lll}
\toprule
~                                                                       & pushpull                 & movetopos               \\
\midrule
Correct single-image action predictions                             &  71.4\%                  &  94.0\%                 \\
\midrule
Successful trajectories with a correct first-state prediction       &  96.4\%                  &  94.0\%                 \\
Successful trajectories with an incorrect first-state prediction    &   3.6\%                  &   6.0\%                 \\
Unsuccessful trajectories with a correct first-state prediction     &   0.0\%                  &   0.0\%                 \\
Unsuccessful trajectories with an incorrect first-state prediction  & 100.0\%                  &   0.0\%                 \\

\bottomrule
\end{tabular}
\end{table*}

Based on our observation that failed trajectories went wrong at the first state, we report how making a correct prediction in the first state relates to the success of that trajectory in Table \ref{results_table_2}. We find that only a small number of the successful trajectories started with an incorrect initial prediction. Secondly, we find that no correct first-state prediction lead to an unsuccessful trajectory. 

\begin{figure}[t]
    \centering
    \includegraphics[width=\linewidth]{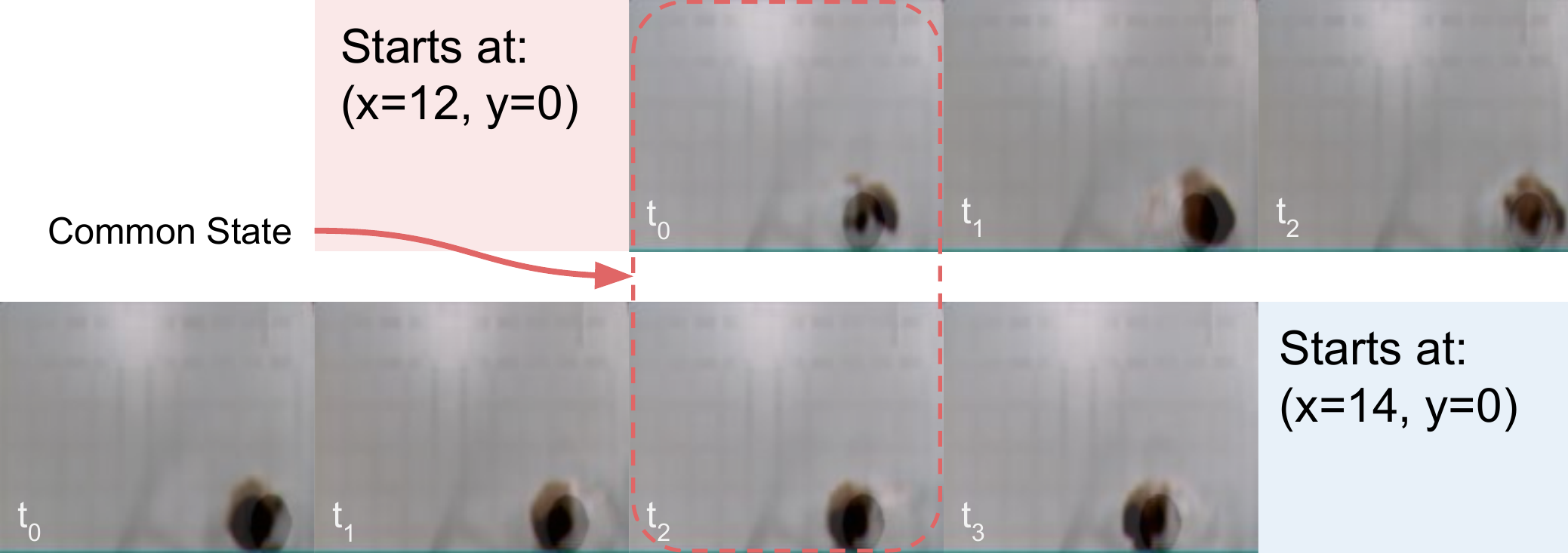}
    \caption{Exemplar of poor first-state prediction compared against a prediction that considers the previous steps. Starting at task-space position (x=12,y=0), an incorrect initial prediction causes the object to move rightwards. A sequence starting at task-space position (x=14, y=0) has built up a correct sequence of actions by the time it arrives at (x=12, y=0) and instead makes a correct prediction and continues to move the object leftwards. Images show the current state with the predicted next state overlaid.}
    \label{fig:single_image_vs_sequence}
\end{figure}

Finally, we highlight an exemplar where a first state, single-image prediction failed but a sequence passing through succeeded. As shown in Figure \ref{fig:single_image_vs_sequence}, the common state of (x=12, y=0) between two different trajectories of the \textit{pushpull} task had two different predictions. While the sequence starting at location (x=12, y=0) failed, the sequence passing through from (x=14, y=0) succeeded in moving through.


\section{Conclusion and Future Work}





We presented a novel methodology for robots to learn tasks from human demonstrations called \textit{Learning from Prediction}. The LfP approach is task-general, robot-general and human-general. These traits are desirable for two key reasons. Firstly, task demonstrations performed by a human do not require knowledge of the target robot. Robot-general demonstrations allow large, freely available video databases such as YouTube to be used. Secondly, on-robot action execution is reduced to the absolute minimum. The robot can perform the task correctly on its first attempt by predicting the outcome of all its actions before choosing to act at every state.

We used the existing PredNet architecture for predicting the outcomes of the human demonstrator and the robot's action primitives. We found PredNet could be successfully trained from only a small number of demonstrations and generalise well to unseen states. For our primary result, we used the sequence of physically visited states to improve prediction performance overall. This lead to 100\% success on the \textit{movetopos} task, accounting for all possible object starting locations.

Using the sequence of previously visited states impacted our desired trait of robot and human generality. In particular, we required that the object movement capabilities of the human and the robot be aligned. We hypothesise that predicting from a single image of the current state alone, will allow for the human and robot to have different object movement capabilities. 

While the number of successful trajectories under single-image prediction was low, we found that only a small percentage of incorrect predictions occurred overall. Under single-image operation, a single incorrect prediction will cause a trajectory failure. Future work will investigate how the single image prediction performance of PredNet can be improved and potential remedies for recovering from an incorrect prediction. 

Future work will also seek to apply the proposed approach to more complicated, three-dimensional tasks with varying backgrounds and distractors. While these domains are not investigated herein, we argue this work be considered a proof of concept of the approach and introduction to a new robot task learning approach we call \textit{Learning from Prediction}.  \balance

\bibliographystyle{IEEEtran.bst}
\bibliography{references.bib}

\end{document}